\documentclass{article}
\usepackage{iclr2027_conference}
\usepackage{times}

\usepackage{amsmath,amsfonts,bm}

\def\eqref#1{equation~\ref{#1}}

\def\1{\bm{1}}

\DeclareMathAlphabet{\mathsfit}{\encodingdefault}{\sfdefault}{m}{sl}
\SetMathAlphabet{\mathsfit}{bold}{\encodingdefault}{\sfdefault}{bx}{n}

\usepackage{graphicx}
\usepackage{xcolor}
\usepackage{flafter}
\usepackage{placeins}
\usepackage{booktabs}
\usepackage{amsmath}
\usepackage{xspace}
\usepackage{microtype}
\usepackage{url}
\usepackage{hyperref}

\newcommand{\model}{\textsc{CNCGEN}\xspace}
\newcommand{\dataset}{\textsc{CNCGEN-Dataset}\xspace}
\title{CNCGEN: A Dataset and Framework\\for Machining Process Planning and\\Toolpath Generation from B-rep Models}
\author{Xiaolei Zhou \And Boyi Lin \And Yuchao Feng \And Jianwei Zheng}
\iclrfinalcopy
\hypersetup{pdftitle={CNCGEN: A Dataset and Framework for Machining Process Planning and Toolpath Generation from B-rep Models},pdfauthor={Xiaolei Zhou, Boyi Lin, Yuchao Feng, Jianwei Zheng}}
\begin{document}
\maketitle
\lhead{Preprint. Under review.}
\begin{abstract}
Learning to generate machining process plans and toolpaths from B-rep CAD requires coupling discrete operation decisions with continuous tool motion as the workpiece evolves. Correctly predicting an operation sequence does not by itself ensure correct material removal, because each toolpath acts on the stock left by preceding cuts. We formulate this problem around persistent manufacturing objects: object identity determines the target of an operation, while the evolving stock state conditions the generation of its toolpath. Based on this formulation, we propose \model, a dataset and learning framework for three-axis machining. \dataset contains approximately 50k geometrically verified synthetic machining flows and 800 held-out real CNC records. Each flow aligns B-rep geometry with object-referenced operations, parameterized toolpaths, intermediate stock states, and verification outcomes, enabling supervision of the correspondence between planning decisions and their geometric effects. \model generates operations and toolpaths for selected objects step by step, updating a compact machining state to guide subsequent predictions. During training, a learned surrogate verifier provides material-removal feedback that links local predictions to their geometric consequences. Experiments on synthetic and held-out real CNC records show that \model improves the resulting workpiece geometry and reduces residual material and overcut compared with adapted CNC generation baselines.
\end{abstract}

\section{Introduction}

CNC machining transforms stock material into a target part through a sequence of cutting operations. A boundary representation (B-rep) describes the desired geometry, while a machining process must also specify operation order, target regions, and tool motion. Learning from paired geometry and machining records offers a way to infer these choices for new parts. However, these choices are coupled: each toolpath acts on a selected target and changes the stock encountered by subsequent operations. The challenge is therefore to link discrete operation decisions to continuous tool motion for the same target, while accounting for their cumulative material-removal effects.

Learning-based methods address a partial perspective of this challenge.
CNC-Net predicts tool radii and paths with iterative stock updates, but follows a prescribed milling-then-drilling schedule rather than selecting the operation type at each step~\citep{yavartanoo2024cncnet}.
In contrast, DeepMS learns operation sequences and associated material-removal volumes from voxelized geometry, but does not generate the toolpaths that realize those volumes~\citep{maqueda2025deepms}.
\citet{lee2026correct} proposes combining LLM-based G-code generation with Separation Logic verification, using collision feedback to guide program correction.
However, agreement with an operation sequence does not ensure that the associated toolpaths remove the intended material from the evolving stock.
Joint learning must therefore connect each operation and toolpath to the same manufacturing target and use the resulting geometric effects to guide both subsequent predictions and training.

\begin{figure}[htbp]
        \centering
        \includegraphics[width=\textwidth,trim=4pt 4pt 4pt 4pt,clip]{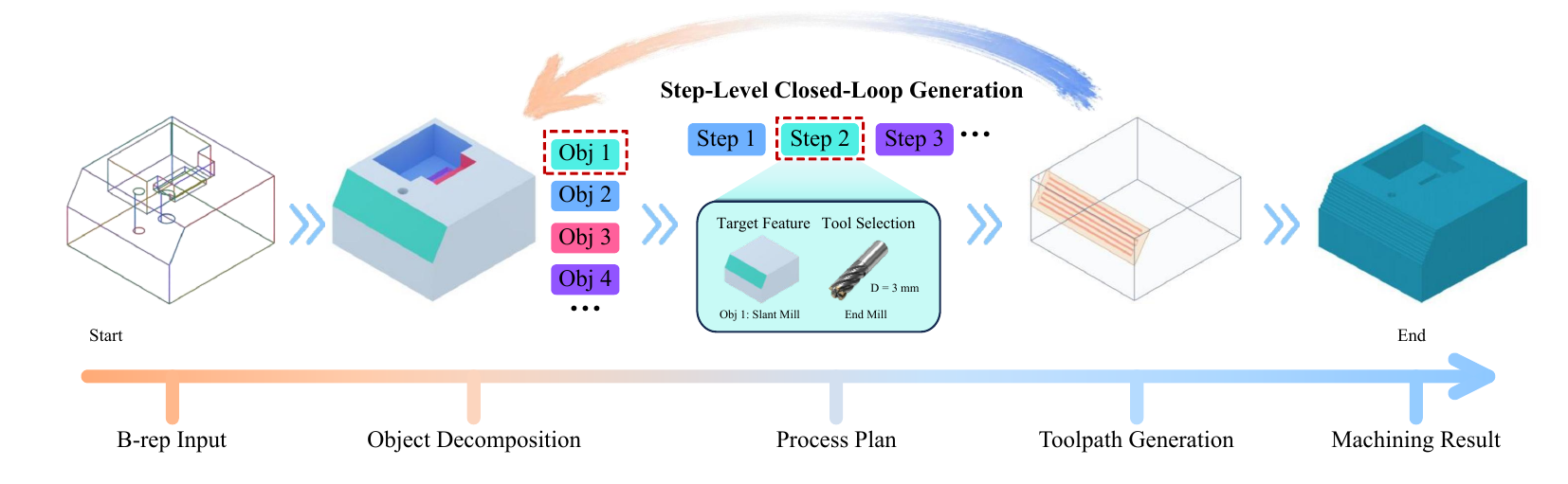}
        \caption{Overview of \model. Given a B-rep model, \model identifies manufacturing objects and generates a machining plan with corresponding toolpaths. The highlighted objects and steps illustrate how a shared identity links discrete operation decisions to continuous tool motion. The upper loop represents feedback of the updated state to subsequent operations and toolpath predictions.}
        \label{fig:teaser}
    \end{figure}

This dependence on machining history is evident when features interact. Consider two pockets whose removal volumes overlap within the same workpiece. Machining one pocket changes the material remaining in the other, although both retain their identities as planning targets. The operation type and target identity alone therefore do not specify the remaining removal task. Subsequent operation and toolpath predictions must account for this changing task while preserving their reference to the same target.
We therefore formulate machining-flow generation as an object-state rollout. Persistent object references link each operation to its target and toolpath, while the evolving machining state conditions subsequent decisions on preceding material removal. Based on this formulation, we introduce \model, a dataset and learning framework for three-axis machining of pockets, holes, chamfers, and slant features, as shown in Fig.~\ref{fig:teaser}.

To supervise this rollout, we construct \dataset with approximately 50k geometrically verified synthetic machining flows. Within each flow, shared object references align operation sequences and parameterized toolpaths with B-rep targets, intermediate stock states, and verification outcomes, so that each operation and toolpath can be related to the stock change it produces. The dataset also includes 800 independently collected, expert-verified real CNC records reserved for evaluating transfer from synthetic to real machining data.

Building on this supervision, \model recovers manufacturing objects from B-rep and stores their geometry, semantics, and evolving states in shared memory. The planner and toolpath decoder read the same indexed objects, giving operation selection and tool motion a common target representation. To account for preceding cuts, a prefix refiner revises proposed operations using the executed prefix and updated state, which also conditions toolpath and cutting-parameter prediction. The estimated removal effects of each generated toolpath update the state for subsequent steps. During training, a surrogate verifier learned from recorded verification outcomes provides differentiable geometric feedback, so that learning accounts for material-removal quality as well as operation and toolpath annotations. Experiments on synthetic and held-out real CNC records show that \model improves material-removal accuracy over CNC-Net with predicted region priors. Ablations further show that removing toolpath state conditioning sharply degrades material-removal accuracy despite nearly unchanged operation-sequence accuracy.
Our overall contributions are summarized as follows.
\begin{itemize}
    \item We formulate joint machining process planning and toolpath generation from B-rep as an object-state rollout, linking discrete operation decisions and continuous tool motion through persistent manufacturing-object identities and evolving machining states.
    \item We construct \dataset with approximately 50k geometrically verified synthetic machining flows and 800 held-out real CNC records. Object-referenced annotations connect operation choices and tool motion to stepwise material-removal effects.
    \item We develop \model, which uses shared object memory to provide a common target representation for planning and toolpath generation, and state-aware prefix refinement and decoding to adapt predictions to preceding cuts. Verification-guided training supplements operation and toolpath supervision with material-removal feedback.
\end{itemize}

    \section{Related Work}

\subsection{B-rep Learning and Manufacturing Feature Recognition}

UV-Net~\citep{jayaraman2021uvnet} encodes geometry in the UV domain and
aggregates features over an adjacency graph, while
BRepNet~\citep{lambourne2021brepnet} defines local convolutions through oriented
coedges. B-rep representation learning also uses contrastive
pretraining~\citep{lou2023brepbert}, masked geometric
reconstruction~\citep{yao2026brepmae,li2026maskedbrep}, and rotation-invariant
geometric context~\citep{ballegeer2026fovnet}. For manufacturing-feature
recognition, Hierarchical CADNet, BRepGAT, and AAGNet identify machining
features from B-rep models~\citep{colligan2022hierarchicalcadnet,lee2023brepgat,wu2024aagnet}.
In \model, the recovered manufacturing objects also serve as persistent
references for operation planning and toolpath generation as the stock evolves.

\subsection{Machining Process Planning and Toolpath Generation}

Feature-based CAPP and STEP-NC systems link geometry, process plans, and
machining instructions through manufacturing knowledge and structured
process descriptions~\citep{nassehi2006stepnc,brecher2006closedloop}.
In commercial CAM, Autodesk Fusion's rest machining restricts toolpaths to
material left by earlier tools or operations~\citep{autodeskFusionRestMachining}.

Learning-based methods infer process routes and operation sequences from
geometric and manufacturing information~\citep{han2023ncplanning,wang2024routeplanning,zhang2024drlroute}.
DeepMS predicts operation sequences and associated material-removal volumes
from voxelized final-part geometry using learned intermediate-state
representations, but does not generate the continuous toolpaths that realize
those volumes~\citep{maqueda2025deepms}. For continuous motion, neural
B-spline surface reparameterization supports real-time toolpath
generation~\citep{feng2024realtime}. CNC-Net~\citep{yavartanoo2024cncnet}
learns tool radii and parameterized milling and drilling paths through
self-supervised shape reconstruction, updating the stock after each operation.
It follows a prescribed milling-then-drilling schedule rather than learning
the operation type at each step.

\citet{lee2026correct} proposes combining LLM-based G-code generation with
STEP-derived geometric constraints and Separation Logic verification.
Collision feedback guides iterative program correction, whereas \model uses
geometric feedback during training to optimize material-removal quality. Its object-state rollout jointly
predicts object-referenced operations and continuous toolpaths, conditioning
subsequent steps on the effects of preceding cuts.

\subsection{Datasets for CAD-to-CAM Learning}

Geometry and design datasets include ABC, which supplies CAD models with
explicit geometric information~\citep{koch2019abc}, and Fusion 360 Gallery,
which records human design sequences~\citep{willis2021fusion360}.
For manufacturing-feature supervision, MFCAD++ provides B-rep
annotations~\citep{colligan2022mfcadplusplus}, while HybridCAD++ extends
feature annotations to hybrid additive and subtractive
manufacturing~\citep{khan2025hybridcadplusplus}. At the process level,
DeepMS associates operation labels with intermediate part geometry and
material-removal volumes~\citep{maqueda2025deepms}.

\dataset explicitly links each object-referenced operation to its
parameterized toolpath and the stock change it produces. B-rep targets and
recorded verification outcomes connect this stepwise supervision to geometric
evaluation, making it possible to assess both the predicted process and its
cumulative material-removal effects.

    \section{Data Representation and Verification}
\label{sec:dataset}

\subsection{Machining Flow Representation}
Each sample $i$ in \dataset is a geometrically verified machining flow represented by
\begin{equation}
\label{eq:flow-record}
d_i=(\mathcal{G}_i,\Omega_{0,i},\Omega_i^\star,\mathcal{O}_i^\star,\mathcal{S}_i^\star,\mathcal{P}_i^\star,\mathcal{X}_i^\star,\mathcal{V}_i).
\end{equation}
Here, $\mathcal{G}_i$ is the input B-rep graph, $\Omega_{0,i}$ the initial stock, and $\Omega_i^\star$ the target part geometry. The remaining fields contain manufacturing objects $\mathcal{O}_i^\star$, operation sequence $\mathcal{S}_i^\star$, corresponding toolpaths $\mathcal{P}_i^\star$, machining-state trajectory $\mathcal{X}_i^\star$, and verification results $\mathcal{V}_i$. Superscript $\star$ is a reference annotation.

Persistent object references link each operation and its toolpath to the same geometric target as the stock changes. The state trajectory records the initial machining state and its evolution after each step, providing the context for successive operations. Appendix~\ref{app:file-format} details the record fields, learning targets, and machining primitives.

\subsection{Dataset Construction and Verification}
\begin{figure}[!htb]
        \centering
        \includegraphics[width=\textwidth]{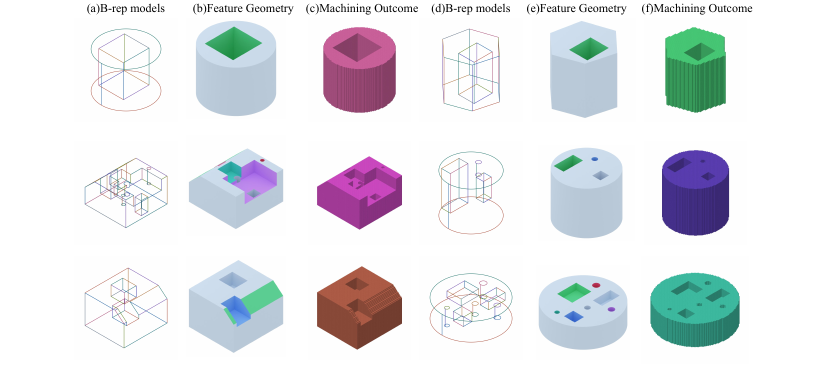}
        \caption{\dataset samples. Wireframe B-rep views are paired with
            rendered machining-feature geometry and verified machining outcomes.
            Fig.~\ref{fig:supp-dataset-gallery} provides additional examples.}
        \label{fig:dataset-presentation}
    \end{figure}

\dataset contains approximately 50k geometrically verified synthetic machining flows and 800 held-out real CNC records. We generate synthetic candidates from standard-part families under constrained three-axis milling rules, covering pockets, holes, chamfers, and slant features. Fig.~\ref{fig:dataset-presentation} shows representative B-rep models, feature geometry, and machining outcomes.

Candidate flows are retained after checks of B-rep geometry, manufacturing objects, and toolpath legality. Material-removal verification compares the resulting stock with the target geometry and records the geometric outcomes of the machining flow.

The real CNC records are independently collected from actual machining scenarios, screened through expert review, and converted to the same flow representation. All 800 records are reserved for testing and excluded from training, validation, model selection, and fine-tuning. This benchmark measures transfer beyond the generated standard-part distribution. Appendix~\ref{app:dataset-details} provides generation parameters, filtering criteria, collection protocols, and dataset statistics.

\section{Methodology}
\label{sec:model}

\begin{figure}[!t]
    \centering
    \includegraphics[width=\textwidth,trim=2pt 2pt 2pt 2pt,clip]{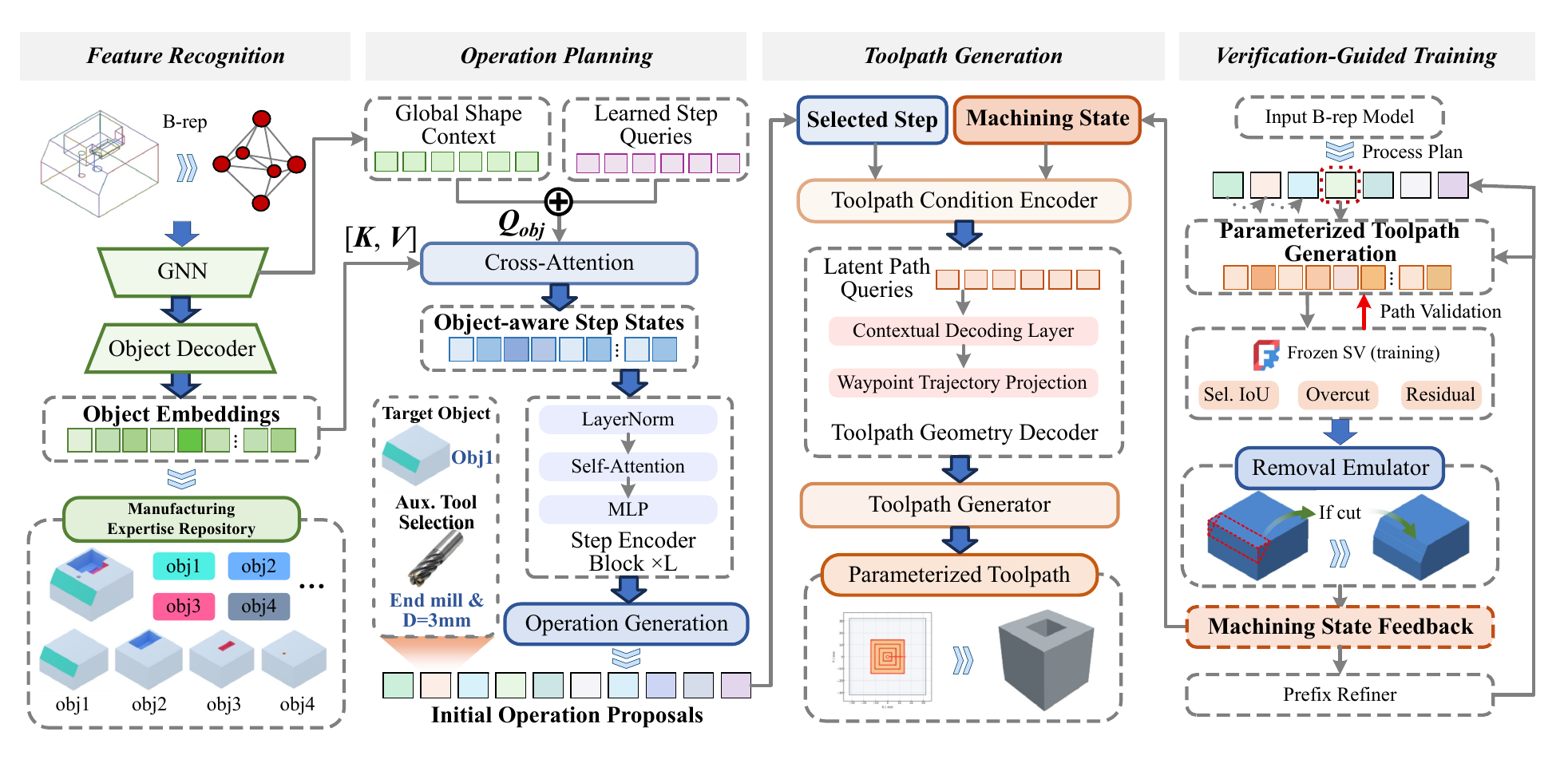}
    \caption{Architecture of \model. The four panels show feature recognition, operation planning, toolpath generation, and verification-guided training. Machining-state feedback links toolpath generation with prefix refinement. The Surrogate Verifier is frozen during generator refinement.}
    \label{fig:method-architecture}
\end{figure}

Recall that operation planning and toolpath generation must share machining targets while adapting to evolving stock. \model couples them through persistent object memory and state feedback (Fig.~\ref{fig:method-architecture}). A parallel planner proposes operations, a prefix refiner revises them using previous decisions and predicted material removal, and state-conditioned toolpaths drive subsequent updates. A frozen Surrogate Verifier (SV) supplies geometric feedback during training.

\subsection{Object Memory from B-rep Geometry}

To give operation planning and toolpath generation a common target reference, we recover persistent manufacturing objects from the B-rep. We represent the input B-rep as $\mathcal{G}=(\mathcal{F},\mathcal{E},\mathbf{X}_{f},\mathbf{X}_{e},\mathbf{g}_{\mathrm{in}})$. The sets $\mathcal{F}$ and $\mathcal{E}$ specify faces and their adjacencies, while $\mathbf{X}_{f}$, $\mathbf{X}_{e}$, and $\mathbf{g}_{\mathrm{in}}$ contain face, edge, and global shape attributes, respectively. A B-rep encoder produces face embeddings $\mathbf{H}$ and a global token $\mathbf{g}$. Following set prediction and object-centric slot modeling~\citep{lee2019settransformer,locatello2020slotattention}, an object decoder uses learned queries to predict up to $K$ object candidates $o_i=(\mathbf{f}_i,\rho_i,\mathbf{z}_i,c_i)$ from these features. Here, $\mathbf{f}_i$ contains manufacturing semantics and geometry, $\rho_i$ associates the object with its B-rep face group, $\mathbf{z}_i$ is its embedding, and $c_i$ is its confidence. During training, Hungarian matching aligns candidate slots with the reference objects in $\mathcal{O}^{\star}$.

The Manufacturing Expertise Repository (MER) retains slots with $c_i>\theta_{\mathrm{obj}}$. For the retained index set $\mathcal{I}$, its entries at step $t$ are
\begin{equation}
\label{eq:mer-entry}
\mathcal{O}_t=\{e_{i,t}\}_{i\in\mathcal{I}},
\qquad
e_{i,t}=(\rho_i,\mathbf{f}_i,\mathbf{z}_i,\mathbf{h}_{i,t},c_i).
\end{equation}
Each MER entry combines fixed object fields $(\rho_i,\mathbf{f}_i,\mathbf{z}_i,c_i)$ with a dynamic token $\mathbf{h}_{i,t}$. The fixed fields preserve target identity throughout the rollout, while $\mathbf{h}_{i,t}$ captures the object's evolving machining context. The compact state $x_t=(\mathbf{m}_t,\mathbf{M}_t,\mathbf{u}_t,\mathbf{U}_t)$ contains a global vector and grid $(\mathbf{m}_t,\mathbf{M}_t)$ summarizing material removal, together with object-level progress variables $(\mathbf{u}_t,\mathbf{U}_t)$. The token combines the fixed features $(\mathbf{f}_i,\mathbf{z}_i)$ with the global state and the state components associated with object $i$:
\begin{equation}
\label{eq:mer-state-token}
\mathbf{h}_{i,t}=\psi_{\theta}\!\left(
[\mathbf{f}_i;\mathbf{z}_i;\mathbf{m}_t;\operatorname{pool}(\mathbf{M}_t);
\mathbf{u}_{i,t};\operatorname{pool}(\mathbf{U}_{i,t})]\right).
\end{equation}
Here, semicolons denote concatenation and $\operatorname{pool}$ aggregates grid features. Both decoders attend to the same indexed entries. Their target references remain fixed as state updates change the context used to predict each operation and its toolpath. Appendices~\ref{app:memory-readout} and~\ref{app:object-objective} give the attention equations and object-supervision loss.

\subsection{Operation Planning with State Feedback}
\label{subsec:object-state-rollout}

Parallel proposals establish an initial sequence, but later decisions must reflect the stock produced by preceding toolpaths. We therefore combine parallel planning with state-dependent prefix refinement. Learned step queries attend to the recovered objects to produce proposals $\tilde{\mathcal{S}}=\{\tilde{s}_t\}_{t=1}^{T}$. Each operation is $s_t=(a_t,\phi_t,\pi_t,\nu_t,\sigma_t)$: operation type, feature type, object pointer, validity, and termination. The pointer $\pi_t$ selects a retained MER entry. An auxiliary coarse tool-selection prediction provides conditioning, while the final tool class is predicted with the toolpath attributes.

At step $t$, the prefix refiner combines the proposal with the current rollout context:
\begin{equation}
\label{eq:prefix-refinement}
s_t=R_{\theta}(\tilde{s}_t,\mathbf{c}_{t-1},x_{t-1},\mathcal{O}_{t-1},\mathcal{G}).
\end{equation}
Here, $R_{\theta}$ denotes refinement and step decoding, and $\mathbf{c}_{t-1}$ summarizes previous decisions, object usage, and path statistics. The refined operation selects the toolpath target, and its predicted material-removal effects update the context for the next step.

Planner supervision uses the operation labels and matched object pointers in $\mathcal{S}^{\star}$, together with coverage, count, and compatibility constraints. Appendix~\ref{app:planner-objective} specifies the pointer distribution and supervised planning objective.

\subsection{State-Conditioned Toolpaths and State Updates}

The same operation on the same object can require different tool motion after preceding cuts. The toolpath decoder therefore conditions each valid step on the selected object, operation, B-rep context, and state $x_{t-1}$. It outputs $\tau_t=(C_t,W_t,\eta_t,\kappa_t,\xi_t)$: segmented cubic B\'ezier control points $C_t$, sampled waypoints $W_t$, pointwise motion and pass labels $\eta_t$, machining strategy and tool class $\kappa_t$, and cutting parameters $\xi_t$. The latter include tool diameter, feedrate, plunge rate, and spindle speed. Corresponding records in $\mathcal{P}^{\star}$ supervise these outputs.

The state transition first estimates the geometric effect of the generated toolpath and then applies a learned correction:
\begin{equation}
\label{eq:state_update}
\begin{aligned}
\bar{x}_t &= T(x_{t-1},s_t,\tau_t),\\
x_t &= \operatorname{clip}\!\left(
\bar{x}_t+
U_{\psi}(\bar{x}_t,o_{\pi_t},\mathbf{e}(s_t),\operatorname{stat}(\tau_t)),0,1
\right).
\end{aligned}
\end{equation}
The deterministic transition $T$ uses compact 2.5D carving to estimate removal depth and coverage on global and object-local grids. This geometric estimate provides $\bar{x}_t$, while $U_{\psi}$ learns a correction from the selected object, step embedding $\mathbf{e}(s_t)$, and toolpath statistics $\operatorname{stat}(\tau_t)$. The corrected state refreshes MER tokens for subsequent predictions. Recorded states in $\mathcal{X}^{\star}$ supervise the update where available. Appendices~\ref{app:toolpath-objective} and~\ref{app:toolpath-state-details} describe B\'ezier sampling, decoding, and the state transition.

\subsection{Verification-Guided Training}
\label{subsec:surrogate-verifier}
\label{subsec:joint-refinement}

Matching operation labels and reference paths supervises local predictions but does not directly score their material-removal consequences. We introduce the SV to turn offline verification into differentiable geometric feedback for each candidate step:
\begin{equation}
\hat{\mathbf{y}}^{\mathrm{sv}}_t
=V_{\omega}(x_{t-1},e_{\pi_t,t-1},s_t,\tau_t).
\end{equation}
The SV combines analytic path-contact features with a learned residual calibration head, pretrained using offline material-removal records $\mathcal{V}$. Its outputs include selected-region IoU, overcut, residual material, aircut, validity, and risk. During refinement, the SV remains frozen while gradients through its differentiable outputs update the generator.

The geometric penalty averages predicted selected-region error over valid candidate steps:
\begin{equation}
\label{eq:sv_loss}
\mathcal{L}_{\mathrm{sv}}=
\frac{1}{|\mathcal{T}_{\mathrm{eval}}|}
\sum_{t\in\mathcal{T}_{\mathrm{eval}}}
\left(1-\widehat{\operatorname{IoU}}_{\mathrm{sel},t}
+0.5\,\hat{p}_{\mathrm{overcut},t}^{\mathrm{sel}}
+0.5\,\hat{p}_{\mathrm{residual},t}^{\mathrm{sel}}\right).
\end{equation}
Hats denote SV predictions, and the $p$ terms are bounded penalties for overcut and residual material. The set $\mathcal{T}_{\mathrm{eval}}$ contains steps with generated toolpaths that pass both planner and path-validity masks. Samples without such steps contribute no SV penalty. The loss combines selected-region IoU error with overcut and residual penalties, while the SV's aircut, validity, and risk outputs serve as auxiliary diagnostics.

Staged pretraining is followed by joint refinement on the generator's own rollouts. The objective combines object, planning, toolpath, state, and verification terms:
\begin{equation}
\label{eq:joint-objective}
\begin{aligned}
\mathcal{L}_{\mathrm{joint}}
&=\lambda_{\mathrm{obj}}\mathcal{L}_{\mathrm{obj}}
+\lambda_{\mathrm{plan}}\mathcal{L}_{\mathrm{plan}}
+\lambda_{\mathrm{path}}\mathcal{L}_{\mathrm{path}}\\
&\quad+\lambda_{\mathrm{state}}\sum_t\|x_t-x_t^\star\|_1
+\lambda_{\mathrm{sv}}\mathcal{L}_{\mathrm{sv}}.
\end{aligned}
\end{equation}
Reference annotations supervise the object, planning, and toolpath losses. The state loss compares the predicted $x_t$ with $x_t^\star$, recorded after the corresponding reference prefix. Appendices~\ref{app:method-objective-details}, \ref{app:training-details}, and~\ref{app:sv-reliability} provide component losses, the training schedule, and SV reliability checks, respectively.

At inference, object recovery and parallel planning provide the object memory and initial operation sequence. Prefix refinement, toolpath generation, and state updates then proceed step by step to produce the machining flow. The SV supplies training feedback and offline diagnostics; reported geometry metrics are computed by the offline evaluator.

\section{Experiments and Analysis}
\label{sec:experiments}

We evaluate whether coupling operation decisions with state-conditioned tool motion improves material removal, which components contribute to this behavior, and how well synthetic training transfers to held-out real CNC records.

\subsection{Experimental Setup}
\label{subsec:experimental-setup}

We split the approximately 50k synthetic machining flows deterministically into 90/5/5 training/validation/test sets. The 800 real CNC records are held out exclusively for testing. All methods share B-rep preprocessing, stock coordinates, millimeter units, target construction, and an offline near-net-stock evaluator with a 4.0\,mm occupancy grid and 8192 sampled surface points. Failed executions remain in evaluation: partial rollouts use their last exported occupancy, and missing outputs are scored as uncut stock. Appendices~\ref{app:unified-geometry-evaluator}, \ref{app:training-details}, and~\ref{app:real-data-target-construction} detail the evaluator, training settings, and real-data targets.

\subsection{Evaluation Metrics}
\label{subsec:evaluation-metrics}

Let $S$ denote the initial stock occupancy, $T$ the target occupancy, and $\hat{T}$ the final occupancy produced by a generated machining result. The target and predicted removal volumes are $R=S\setminus T$ and $\hat{R}=S\setminus\hat{T}$. Final-shape IoU measures agreement between $\hat{T}$ and $T$, while Removal F1 combines removal precision and recall over $\hat{R}$ and $R$. Chamfer Distance (CD), normalized by the stock bounding-box diagonal, measures surface discrepancy. Overcut $|\hat{R}\cap T|/|T|$ measures removal of material that should remain, and residual $|\hat{T}\cap R|/|R|$ measures material left in the intended removal region. Together, these metrics assess final geometry and the extent and accuracy of material removal.

For \model ablations, step and sequence exact match compare operation types, feature types, and object references at the corresponding levels. These scores apply to variants with stable object references. Category macro-F1 summarizes categorical prediction quality.

\subsection{Baselines and Comparison Setting}

We adapt CNC-Net~\citep{yavartanoo2024cncnet} to the B-rep input setting with a Machining Region Prior (MRP), which supplies a machining-region support mask and a depth field to its carving model. CNC-Net+Pred.\ MRP serves as the main learned baseline. A region head trained on the training split predicts the prior from the input B-rep, so test-time generation uses the input geometry. CNC-Net+Oracle MRP uses priors derived from ground-truth manufacturing-feature and toolpath annotations as a diagnostic of process generation under accurate region localization.

Both variants are trained from scratch on the same splits as \model and use the same validation protocol and geometry evaluator. Each variant is trained and evaluated with its corresponding prior type. Under the shared evaluation, the learning formulations differ: CNC-Net consumes rasterized support and depth fields, whereas \model uses B-rep graph features with object, operation, toolpath, and state supervision. Appendix~\ref{app:cncnet-mrp-variants} details MRP construction, rasterization, and the CNC-Net adaptation.

\subsection{Main Comparison}

On the synthetic test set, \model substantially improves material-removal quality over CNC-Net+Pred.\ MRP (Table~\ref{tab:verification-quality}). Removal F1 increases from 0.2318 to 0.9016, while Residual decreases from 0.6032 to 0.0431, a relative reduction of 92.9\%. IoU also increases from 0.7144 to 0.9525, with lower CD and Overcut. Together, these results show more complete removal of the intended material and less unintended removal of the target part.

Oracle region priors improve the adapted CNC-Net baseline but leave substantial residual material. CNC-Net+Oracle MRP achieves an IoU of 0.9271 and an Overcut of 0.0025, yet its Residual remains 0.5973. Compared with this variant, \model reduces Residual to 0.0431 and increases Removal F1 from 0.5067 to 0.9016, although its Overcut is higher (0.0079 versus 0.0025). Thus, accurate region localization alone does not ensure complete material removal in this baseline. Removal F1 and Residual reveal differences in machining completeness that final-shape IoU alone does not fully capture. Appendix~\ref{app:failure-case-analysis} presents representative failure cases.

\begin{table}[!htbp]
    \centering
    \caption{Synthetic-test geometry quality after execution. CNC-Net+Pred.\ MRP uses predicted region priors; Oracle MRP uses annotation-derived priors as a diagnostic setting. CD is normalized by the stock bounding-box diagonal. Bold marks the best value in each column.}
    \label{tab:verification-quality}
    \scriptsize
    \setlength{\tabcolsep}{3pt}
    \resizebox{\columnwidth}{!}{%
        \begin{tabular}{@{}llccccc@{}}
            \toprule
            Method & MRP source & IoU $\uparrow$ & Removal F1 $\uparrow$ & CD $\downarrow$ & Overcut $\downarrow$ & Residual $\downarrow$ \\
            \midrule
            CNC-Net+Pred. MRP & Predicted & 0.7144 & 0.2318 & 0.0595 & 0.2327 & 0.6032 \\
            CNC-Net+Oracle MRP & Oracle & 0.9271 & 0.5067 & 0.0309 & \textbf{0.0025} & 0.5973 \\
            \textbf{\model} (Ours) & N/A & \textbf{0.9525} & \textbf{0.9016} & \textbf{0.0268} & 0.0079 & \textbf{0.0431} \\
            \bottomrule
        \end{tabular}
    }
\end{table}

\subsection{Ablation Studies}

The ablations examine target consistency, cross-step correction, state-conditioned motion, and geometric training feedback by individually removing MER, the prefix refiner, state-aware toolpaths, and the SV loss (Table~\ref{tab:ablation}). All variants use the same data split, training protocol, and evaluation interface.

\begin{table}[!htbp]
    \centering
    \caption{Ablation results on \dataset. Each variant disables one component of \model while preserving the same evaluation interface. Bold marks the best value in each column.}
    \label{tab:ablation}
    \scriptsize
    \setlength{\tabcolsep}{1.5pt}
    \resizebox{\columnwidth}{!}{%
        \begin{tabular}{@{}lcccccccc@{}}
            \toprule
            Variant & Step EM $\uparrow$ & Seq. EM $\uparrow$ & Macro-F1 $\uparrow$ & IoU $\uparrow$ & Removal F1 $\uparrow$ & CD $\downarrow$ & Overcut $\downarrow$ & Residual $\downarrow$ \\
            \midrule
            \textbf{Full \model} (Ours) & \textbf{0.8552} & \textbf{0.8290} & \textbf{0.9024} & \textbf{0.9525} & \textbf{0.9016} & \textbf{0.0268} & \textbf{0.0079} & \textbf{0.0431} \\
            w/o MER & N/A$^\dagger$ & N/A$^\dagger$ & 0.2182 & 0.9315 & 0.1484 & 0.0317 & 0.0086 & 0.0673 \\
            w/o prefix refiner & 0.8172 & 0.7796 & 0.8954 & 0.9332 & 0.3115 & 0.0274 & 0.0167 & 0.0475 \\
            w/o state-aware toolpaths & 0.8530 & 0.8260 & 0.9012 & 0.9327 & 0.3113 & 0.0273 & 0.0169 & 0.0477 \\
            w/o SV loss & 0.8400 & 0.8132 & 0.8778 & 0.9296 & 0.3041 & 0.0286 & 0.0236 & 0.0472 \\
            \bottomrule
        \end{tabular}
    }
    \parbox{\columnwidth}{\scriptsize $^\dagger$Step and sequence exact match are not applicable because removing the MER removes stable operation-target identities; the variant is therefore evaluated with category and executed-geometry metrics.}
\end{table}

State conditioning has a much larger effect on material-removal quality than on operation-sequence accuracy. Without state-aware toolpaths, Seq.\ EM changes from 0.8290 to 0.8260, while Removal F1 falls from 0.9016 to 0.3113 and overcut rises from 0.0079 to 0.0169. Thus, similar sequence accuracy can accompany very different removal outcomes, supporting the use of the evolving machining state to condition tool motion.

Removing MER reduces category Macro-F1 from 0.9024 to 0.2182 and Removal F1 from 0.9016 to 0.1484. Removing the prefix refiner lowers Seq.\ EM from 0.8290 to 0.7796 and Removal F1 to 0.3115. These results support the contributions of persistent object memory and cross-step refinement to coupling operation decisions with material removal.

Without the SV loss, IoU decreases from 0.9525 to 0.9296 and overcut increases from 0.0079 to 0.0236, the highest among the ablations. Held-out SV diagnostics assess the accuracy of the feedback model (Appendix~\ref{app:sv-reliability}), whereas these generator results use the offline geometry evaluator. Appendix~\ref{app:detailed-ablation-analysis} provides detailed component analysis.

    \subsection{Real-world Evaluation}
\label{subsec:real-world-evaluation}

\begin{figure}[htbp]
    \centering
    \includegraphics[width=\textwidth,height=0.36\textheight,keepaspectratio,trim=6pt 6pt 6pt 6pt,clip]{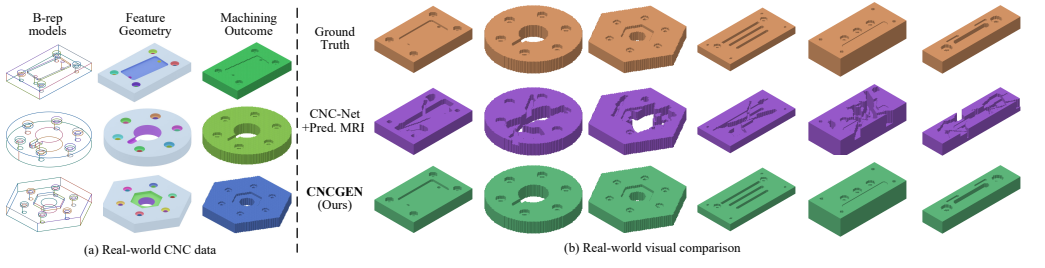}
    \caption{Geometric outcomes on held-out real CNC records. Left: representative B-rep models, feature geometry, and machining outcomes from the benchmark. Right: target geometry and final geometries produced by \model and CNC-Net+Pred.\ MRP under the offline evaluator.}
    \label{fig:realcompare}
\end{figure}

We evaluate transfer to the 800 independently collected, expert-verified CNC records using the held-out protocol in Section~\ref{subsec:experimental-setup} and the same near-net-stock evaluator as for synthetic data. Targets are derived from verified manufacturing annotations, with auxiliary voxel records used for consistency checks (Appendix~\ref{app:real-data-target-construction}). In the real-part examples shown in Fig.~\ref{fig:realcompare}, \model recovers cavities and through-regions more completely, whereas CNC-Net+Pred.\ MRP leaves residual structures within these regions. This qualitative difference is consistent with \model's higher Removal F1 and lower Residual in Table~\ref{tab:realcnc-verification-generalization}.

\begin{table}[!htbp]
    \centering
    \caption{Geometry quality on 800 held-out real CNC records, evaluated with the same near-net-stock protocol as the synthetic test set. Pred.\ MRP uses predicted region priors, while Oracle MRP uses annotation-derived priors. CD is normalized by the stock bounding-box diagonal. Bold marks the best value in each column.}
    \label{tab:realcnc-verification-generalization}
    \scriptsize
    \setlength{\tabcolsep}{3pt}
    \resizebox{\columnwidth}{!}{%
        \begin{tabular}{@{}llccccc@{}}
            \toprule
            Method & MRP source & IoU $\uparrow$ & Removal F1 $\uparrow$ & CD $\downarrow$ & Overcut $\downarrow$ & Residual $\downarrow$ \\
            \midrule
            CNC-Net+Pred. MRP & Predicted & 0.6275 & 0.1170 & 0.0703 & 0.3205 & 0.7268 \\
            CNC-Net+Oracle MRP & Oracle & 0.9203 & 0.4290 & 0.0332 & \textbf{0.0112} & 0.6617 \\
            \textbf{\model} (Ours) & N/A & \textbf{0.9274} & \textbf{0.8126} & \textbf{0.0148} & 0.0134 & \textbf{0.1845} \\
            \bottomrule
        \end{tabular}
    }
\end{table}

\model improves all five geometry metrics over CNC-Net+Pred.\ MRP on this benchmark without fine-tuning (Table~\ref{tab:realcnc-verification-generalization}). Removal F1 increases from 0.1170 to 0.8126, and residual decreases from 0.7268 to 0.1845. The comparison with Oracle MRP follows the same pattern as on synthetic data: \model achieves more complete removal, whereas CNC-Net+Oracle MRP has lower overcut (0.0112 versus 0.0134). Oracle MRP leaves a residual of 0.6617 versus 0.1845 for \model, with a lower Removal F1 of 0.4290 versus 0.8126.

Relative to synthetic-test results, \model's Removal F1 decreases from 0.9016 to 0.8126 and residual increases from 0.0431 to 0.1845. Case inspection identifies remaining errors in multi-feature pockets, narrow chamfers, slant features, and overlapping pocket boundaries.

    \section{Limitations}

\model targets three-axis machining of pockets, holes, chamfers, and slant features. External profile machining is supplied by setup construction, while turning, free-form finishing, and simultaneous multi-axis machining require broader process or motion representations. The compact 2.5D state and 4.0\,mm occupancy evaluation describe material removal at a coarse geometric scale. Narrow-feature and boundary errors remain in the real benchmark, and the reported metrics do not establish machining-tolerance or surface-finish accuracy.

Training uses synthetic standard-part flows, while the 800 real records cover a limited range of shop practices, fixtures, and machining intent. Generated plans and parameterized toolpaths require machine-specific post-processing. Geometric verification leaves cutting forces, tool wear, chatter, and the physical suitability of predicted cutting parameters unassessed. Appendix~\ref{app:limitations-details} discusses further process boundaries and extensions.

\section{Conclusion}

We introduced \dataset and \model for jointly learning object-referenced process plans and continuous toolpaths from B-rep models. Shared target identities and evolving stock states connect operation decisions with tool motion. Experiments on synthetic and held-out real records show improved material-removal quality over CNC-Net with predicted region priors. The ablations show that similar sequence accuracy can mask substantially different removal quality, supporting state-conditioned toolpath generation.

    \clearpage
\section*{AI Use Disclosure}
Generative AI tools assisted manuscript preparation through language editing, translation, structural revision, and editorial feedback on the presentation of methods and results. They were not used to generate the dataset, implement the proposed method, or conduct the reported experiments. The authors take responsibility for the final manuscript and its scientific claims.

    \bibliographystyle{iclr2027_conference}
    \bibliography{sample-bibliography}

\clearpage
\appendix
\section*{Supplementary Material}
\setcounter{figure}{0}
\setcounter{table}{0}
\renewcommand{\thefigure}{S\arabic{figure}}
\renewcommand{\thetable}{S\arabic{table}}
\renewcommand{\theHfigure}{supp.\arabic{figure}}
\renewcommand{\theHtable}{supp.\arabic{table}}
\numberwithin{equation}{section}

This supplement describes dataset construction and the held-out real benchmark
(Section~\ref{app:dataset-details}), model objectives and verifier training
(Section~\ref{app:method-objective-details}), experimental protocols and additional
analyses (Section~\ref{app:evaluation-protocol-details}), and limitations
(Section~\ref{app:limitations-details}). Notation follows the main text.

\section{Dataset Details}
\label{app:dataset-details}

\subsection{Machining Flow Records}
\label{app:file-format}

Each sample in \dataset stores a machining flow under a unique sample identifier,
following the record definition in Equation~\ref{eq:flow-record}. The record
links B-rep geometry $\mathcal{G}_i$, initial stock $\Omega_{0,i}$, and target
part $\Omega_i^\star$ with manufacturing objects, ordered operations, toolpaths,
intermediate states, and verification outcomes. Exported records use millimeter
units and a work coordinate system in which the stock top is at $z=0$.

Object annotations $\mathcal{O}_i^\star$ associate manufacturing semantics and
geometry with B-rep face groups and persistent object references. Operation
records $\mathcal{S}_i^\star$ specify the operation order, operation and feature
types, object pointers, validity labels, and termination labels. Multiple
operations can reference the same object, preserving its identity across
machining steps.

Toolpath records $\mathcal{P}_i^\star$ contain control points and waypoints,
pointwise motion and pass labels, machining strategy, tool class, and cutting
parameters. Object pointers and toolpath identifiers link each operation to
its manufacturing target and corresponding tool motion.

The state trajectory $\mathcal{X}_i^\star=\{x_{i,t}^\star\}_{t=0}^{T_i}$ contains
the initial state and the state after each of the $T_i$ reference machining
steps. Thus, each state corresponds to the material remaining after a specific
executed operation prefix. Verification records $\mathcal{V}_i$ describe the
geometric outcomes of the associated operations and toolpaths. Together, these
records identify what an operation acts on, how the tool moves, and how the
stock changes.

\begin{figure}[htbp]
        \centering
        \includegraphics[
            width=\textwidth,
            height=0.31\textheight,
            keepaspectratio,
            trim=4pt 4pt 4pt 4pt,
            clip
        ]{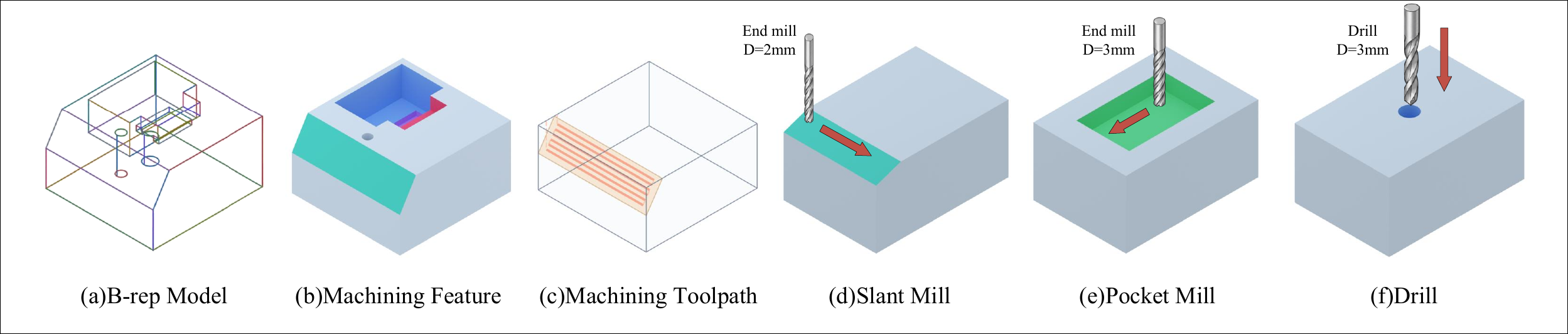}
        \caption{CNC machining primitives and execution
            representations used by \dataset and \model: B-rep input, extracted
            machining feature, recorded toolpath, and operation primitives for
            slant milling, pocket milling, and drilling.}
        \label{fig:cnc-primitives}
    \end{figure}

\begin{figure}[htbp]
    \centering
    \includegraphics[width=\textwidth,height=0.48\textheight,keepaspectratio,trim=24pt 24pt 24pt 24pt,clip]{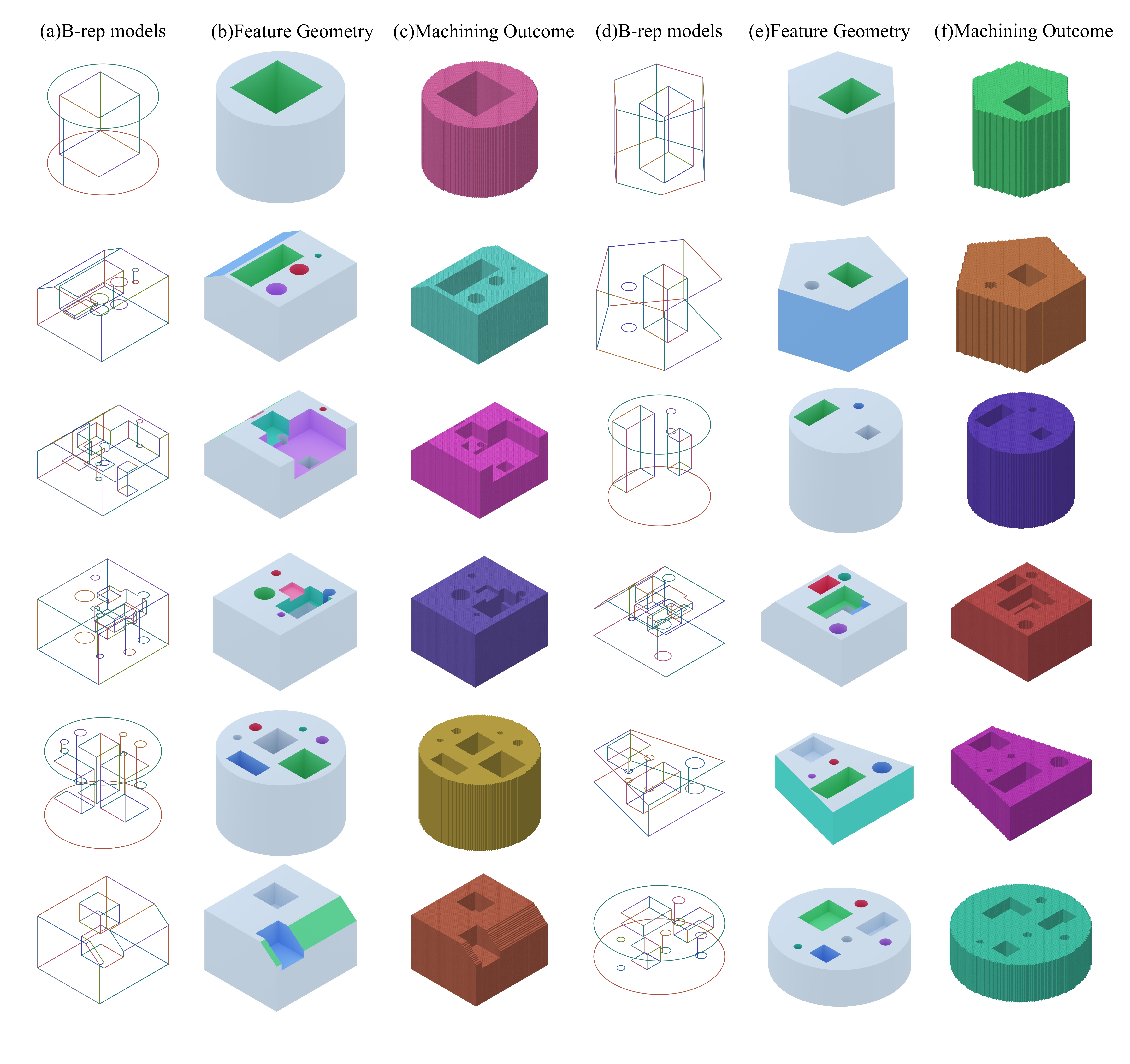}
    \caption{Extended gallery of twelve \dataset examples. Each example pairs
    a wireframe B-rep model with rendered manufacturing-feature geometry and a
    verified machining outcome. Fig.~\ref{fig:dataset-presentation} shows a subset.}
    \label{fig:supp-dataset-gallery}
\end{figure}

Fig.~\ref{fig:cnc-primitives} illustrates the geometric representations and
operation primitives used in these records. Fig.~\ref{fig:dataset-presentation}
and~\ref{fig:supp-dataset-gallery} show representative dataset examples.

\subsection{Synthetic Data Generation and Verification}
\label{app:construction-filters}

Synthetic machining flows are generated from conventional standard-part
families under constrained three-axis milling rules, covering pockets, holes,
chamfers, and slant features. The generator varies stock geometry, feature
positions, depths and combinations, operation order, and canonical toolpath
parameters. Box or cylindrical stocks are sampled within fixed dimension
ranges, with features placed under clearance constraints. Manufacturing
constraints guide part and process construction, as in the procedural
generation used by DeepMS~\citep{maqueda2025deepms}. The held-out real records
are not used as generation templates or sources of perturbed training examples.

Construction proceeds through B-rep screening and graph extraction,
manufacturability filtering, object and process export, toolpath export, and
material-removal verification. Screening rejects invalid B-reps, degenerate
solids, operation counts outside $1$--$16$, feature depths above $300$\,mm,
in-plane feature sizes above $600$\,mm, boundary complexities outside
$3$--$5000$ points, inaccessible three-axis geometry, and illegal toolpaths.

Executable cases with shallow cuts, narrow features, high depth-to-width
ratios, or feature sizes close to the nominal tool diameter retain difficulty
annotations. Toolpath checks record weak bounding-box overlap, large $z$
errors, large $xy$ jumps, and unexpected cutting heights. A sample is retained
only if it passes the hard toolpath-legality checks and achieves a minimum
voxelized verification IoU of $0.95$ across validation runs.

Exported toolpaths include the geometric and motion fields described in
Appendix~\ref{app:file-format}, together with tool diameter, feedrate, plunge
rate, and spindle speed. Material-removal verification compares the machined
stock with the target geometry and records coverage, overcut, residual material,
aircut, validity, IoU, Removal F1, Chamfer Distance, and execution status. The
resulting reports support construction checks and provide geometric supervision.

\subsection{Real-World Data Collection and Curation}
\label{app:real-world-benchmark}

The real benchmark draws on actual three-axis CNC machining records collected
from laboratory platforms, including Haas Mini Mill, Haas VF-2, and Tormach
770M. More than 2,000 candidate records were gathered for screening and expert
review.

Fifty CNC-domain experts, including manufacturing engineers, CNC programming
engineers, and experienced machinists, assessed three-axis suitability,
geometric validity, feature clarity, feature--operation correspondence,
toolpath metadata completeness, machining complexity, and sample diversity.
Records with unsupported process routes or missing target-construction
metadata were excluded.

\begin{figure}[htbp]
    \centering
    \includegraphics[width=0.92\textwidth,height=0.65\textheight,keepaspectratio,trim=24pt 24pt 24pt 24pt,clip]{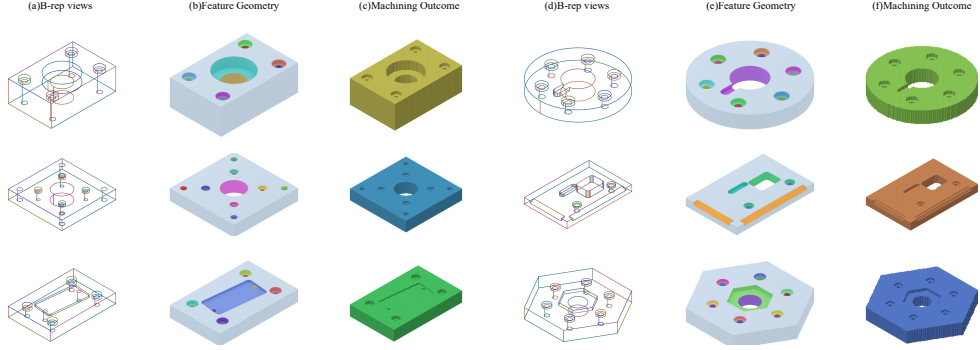}
    \caption{Representative B-rep parts from the held-out real CNC benchmark,
    illustrating machining geometries beyond the synthetic standard-part families.}
    \label{fig:realdata-presentation}
\end{figure}

Screening and expert review retained 800 records with aligned part geometry,
stock information, manufacturing features, operation descriptions, toolpath
metadata, and machining outcomes. These records follow the same machining-flow
interface as the synthetic data. All 800 records are held out for testing and
are excluded from training, validation, model selection, and fine-tuning.
Fig.~\ref{fig:realdata-presentation} shows representative parts.

\subsection{Split Protocol and Dataset Statistics}
\label{app:dataset-statistics}

\dataset contains approximately 50k geometrically verified synthetic machining
flows and 800 held-out real CNC records. The synthetic flows are partitioned
into training, validation, and test sets using a deterministic 90/5/5 split.
Partition membership is determined by hashing the sample identifier. Applying
the same split rule to the same identifier therefore preserves its assignment
regardless of record order.

\begin{figure}[htbp]
    \centering
    \includegraphics[width=0.95\textwidth,trim=2pt 2pt 2pt 0pt,clip]{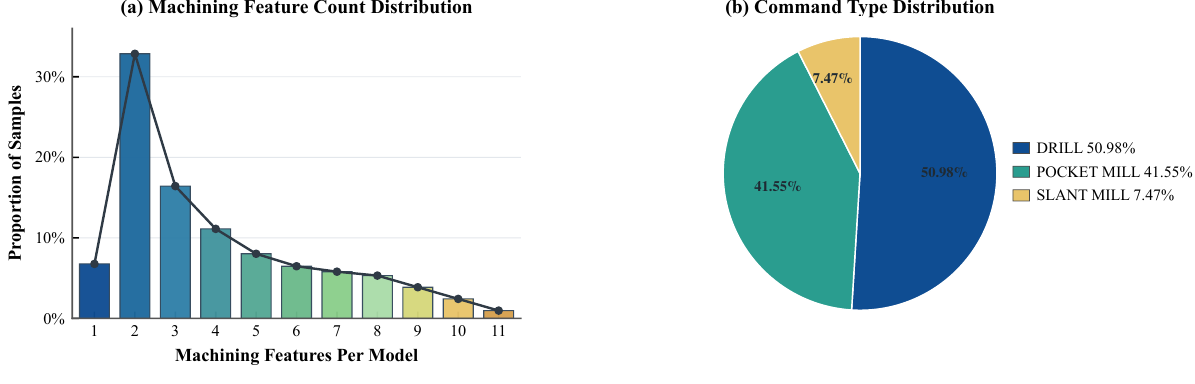}
    \caption{Synthetic dataset statistics: distribution of feature counts and
    proportions of command types in the generated machining flows.}
    \label{fig:supp-dataset-distribution}
\end{figure}

Fig.~\ref{fig:supp-dataset-distribution} summarizes the synthetic dataset through
the distribution of feature counts per flow and the proportions of command types.

\section{Method Details}
\label{app:method-objective-details}

\begin{figure}[!htb]
        \centering
        \includegraphics[width=\textwidth,height=0.31\textheight,keepaspectratio,trim=12pt 12pt 12pt 12pt,clip]{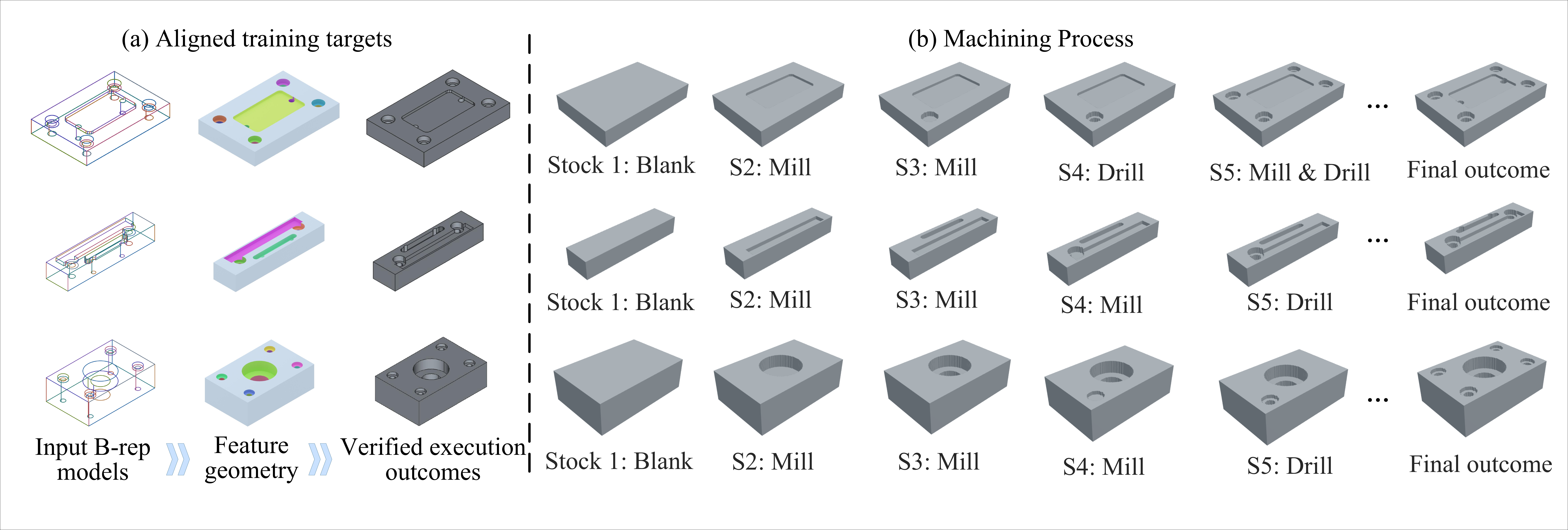}
        \caption{Aligned training targets and verification supervision. Representative
            machining flows align B-rep geometry, machining-feature targets,
            toolpath/process targets, and verified execution outcomes; the verifier
            supervision compares generated stock evolution with the target geometry.}
        \label{fig:process-supervision}
    \end{figure}

The record $d_i$ in Equation~\ref{eq:flow-record} supplies aligned targets for
object recovery, planning, toolpath generation, state prediction, and
verification. The model uses the recorded state after the corresponding
reference prefix for state supervision, and it updates its own predicted state
when generating a rollout. Fig.~\ref{fig:process-supervision} illustrates the
aligned training targets and material-removal verification outcomes.

\subsection{Object Encoding and Shared Memory}
\label{app:memory-readout}

The B-rep encoder $E_\theta$ maps the input geometry $\mathcal{G}$ to face
embeddings $\mathbf{H}$ and a global shape token $\mathbf{g}$. Given these
features, the object decoder $D_\theta$ uses $K$ learned queries to produce
candidate manufacturing objects:
\begin{equation}
\begin{aligned}
(\mathbf{H},\mathbf{g})&=E_\theta(\mathcal{G}),\\
o_i&=D_\theta(\mathbf{q}_i,\mathbf{H},\mathbf{g}),\qquad i=1,\ldots,K.
\end{aligned}
\end{equation}
Here, $\mathbf{q}_i$ denotes the learned query for candidate $i$. Confidence
filtering determines the retained object index set $\mathcal{I}$. Each retained
MER entry combines fixed object fields with a step-dependent machining-context
token $\mathbf{h}_{i,t}$, as defined in Equations~\ref{eq:mer-entry}
and~\ref{eq:mer-state-token} of the main text.

Operation planning and toolpath generation access the same retained MER entries
through query-dependent attention. For a query $\mathbf{r}$, the attention
keys, weights, and memory readout are computed as follows:
\begin{equation}
\begin{aligned}
\mathbf{k}_{i,t}&=\mathbf{W}_k[\operatorname{emb}(\rho_i);\mathbf{f}_i;\mathbf{z}_i;\mathbf{h}_{i,t}],\\
\alpha_i(\mathbf{r},t)&=\operatorname{softmax}_{i\in\mathcal{I}}
\left(\frac{(\mathbf{W}_q\mathbf{r})^\top\mathbf{k}_{i,t}}{\sqrt{d}}\right),\\
\operatorname{read}_{\mathrm{MER}}(\mathbf{r},t)&=\sum_{i\in\mathcal{I}}\alpha_i(\mathbf{r},t)[\mathbf{z}_i;\mathbf{h}_{i,t}].
\end{aligned}
\label{eq:mer-read}
\end{equation}
The function $\operatorname{emb}$ embeds the object reference,
$\mathbf{W}_k$ and $\mathbf{W}_q$ are learned projections, and $d$ is the
key/query dimension. The softmax is normalized over the retained object indices
$\mathcal{I}$. The weighted readout combines object embeddings $\mathbf{z}_i$
with their current machining-context tokens $\mathbf{h}_{i,t}$, allowing both
decoders to access shared object representations with updated state information.

\paragraph{Manufacturing object supervision.}
\label{app:object-objective}

To supervise the predicted object slots, Hungarian assignment establishes
one-to-one correspondences with the reference objects. Let $\mathcal{M}$ denote
the matched index pairs $(i,j)$, where $i$ indexes a predicted slot and $j$
indexes a reference object. The set
$\mathcal{I}_{\mathrm{match}}=\{i:(i,j)\in\mathcal{M}\}$ contains the matched
predicted slots. The object-supervision loss is
\begin{equation}
\begin{aligned}
\mathcal{L}_{\mathrm{obj}}
&=\sum_{(i,j)\in\mathcal{M}}\big[
\lambda_f\lVert\hat{\mathbf{f}}_i-\mathbf{f}_j^\star\rVert_1
+\lambda_\rho\mathcal{L}_{\mathrm{ref}}(\hat{\rho}_i,\rho_j^\star)
+\lambda_c\operatorname{BCE}(\hat{c}_i,1)\big]\\
&\quad+\lambda_{\mathrm{bg}}\sum_{i\notin\mathcal{I}_{\mathrm{match}}}
\operatorname{BCE}(\hat{c}_i,0).
\end{aligned}
\end{equation}
For each matched pair, the $L_1$ term supervises the predicted object fields
$\hat{\mathbf{f}}_i$, while $\mathcal{L}_{\mathrm{ref}}$ supervises associations
with B-rep face groups and persistent object references. The confidence target
is one for matched slots and zero for unmatched slots. The coefficients
$\lambda_f$, $\lambda_\rho$, $\lambda_c$, and $\lambda_{\mathrm{bg}}$ weight the
corresponding loss terms. Hats denote predictions, and $\star$ denotes reference
annotations.

\subsection{Operation Planning and Prefix Refinement}
\label{app:planner-objective}

At step $t$, the planner predicts an object pointer over the retained MER
entries. Given the step query $\mathbf{r}_t$, the pointer distribution is
\begin{equation}
p(\pi_t=i\mid\mathbf{r}_t,\mathcal{O}_{t-1},x_{t-1})
=\operatorname{softmax}_{i\in\mathcal{I}}
\left(\frac{(\mathbf{W}_q\mathbf{r}_t)^\top
\mathbf{W}_k[\mathbf{z}_i;\mathbf{h}_{i,t-1}]}{\sqrt{d}}\right).
\end{equation}
Each pointer logit combines the object's embedding $\mathbf{z}_i$ with its
machining-context token $\mathbf{h}_{i,t-1}$ before the current operation. The
softmax is normalized over the retained object indices $\mathcal{I}$. Reference
object pointers are mapped to predicted slot indices using the Hungarian
assignment described in Section~\ref{app:object-objective}.

Planning supervision covers the operation type $a_t$, feature type $\phi_t$,
object pointer $\pi_t$, validity $\nu_t$, and termination $\sigma_t$.
Cross-entropy supervises the first three outputs, while binary cross-entropy
supervises validity and termination:
\begin{equation}
\begin{aligned}
\mathcal{L}_{\mathrm{plan}}^{\mathrm{sup}}
=\sum_{t=1}^{T}\big[&\operatorname{CE}(\hat a_t,a_t^\star)
+\operatorname{CE}(\hat\phi_t,\phi_t^\star)
+\operatorname{CE}(\hat\pi_t,\pi_t^\star)\\
&+\operatorname{BCE}(\hat\nu_t,\nu_t^\star)
+\operatorname{BCE}(\hat\sigma_t,\sigma_t^\star)\big].
\end{aligned}
\end{equation}
These terms form the supervised component of $\mathcal{L}_{\mathrm{plan}}$;
the planner also uses the coverage, count, and compatibility constraints
described in the main text. During rollout, the prefix refiner conditions each
revised step on previous decisions, object usage, path statistics, and the
updated state through Equation~\ref{eq:prefix-refinement}.

\subsection{Toolpath Generation and State Updates}
\label{app:toolpath-objective}

Each valid machining step is represented by a sequence of cubic B\'ezier
segments. For segment $\ell$ at step $t$, four control points
$C_{t,\ell}=\{\mathbf{c}_{t,\ell,j}\}_{j=0}^{3}$ define the continuous curve:
\begin{equation}
B_{t,\ell}(\alpha)=\sum_{j=0}^{3}\binom{3}{j}
(1-\alpha)^{3-j}\alpha^j\mathbf{c}_{t,\ell,j},\qquad\alpha\in[0,1].
\end{equation}
Evaluating each valid segment at fixed parameter values $\alpha$ and
concatenating the sampled points in segment order produces the waypoint
sequence $W_t$. The toolpath objective supervises both path geometry and
execution attributes. For a valid step $t$, the loss is
\begin{equation}
\begin{aligned}
\ell_{\mathrm{path},t}
&=\lVert C_t-C_t^\star\rVert_1
+\lambda_w\lVert W_t-W_t^\star\rVert_1\\
&\quad+\lambda_\eta\operatorname{CE}(\eta_t,\eta_t^\star)
+\lambda_\kappa\operatorname{CE}(\kappa_t,\kappa_t^\star)
+\lambda_\xi\lVert\xi_t-\xi_t^\star\rVert_1.
\end{aligned}
\end{equation}
The $L_1$ losses on control points $C_t$ and sampled waypoints $W_t$ supervise
path geometry. Cross-entropy supervises the motion and pass labels $\eta_t$
and the strategy and tool classes $\kappa_t$, while an additional $L_1$ term
supervises the scalar cutting parameters $\xi_t$. Reference targets are denoted
by $\star$. The training objective $\mathcal{L}_{\mathrm{path}}$ aggregates these
losses over valid supervised steps.

\paragraph{Decoding and state updates.}
\label{app:toolpath-state-details}

At step $t$, the Toolpath Condition Encoder in
Fig.~\ref{fig:method-architecture} combines the selected operation $s_t$,
target-object embedding, B-rep context, and pre-operation machining state
$x_{t-1}$. Latent path queries attend to this condition through
contextual decoding layers. Waypoint trajectory projection and toolpath
geometry decoding produce B\'ezier control points, sampled waypoints,
pointwise motion and pass labels, strategy/tool attributes, and cutting
parameters.

The deterministic transition $T$ in Equation~\ref{eq:state_update} tracks
material removal with a compact 2.5D carving model. It projects valid sampled
waypoints onto global and object-local material grids, expands their footprints
by the normalized cutter radius, and accumulates removal depth and coverage.
This produces the intermediate state $\bar{x}_t$.

Starting from the geometric estimate $\bar{x}_t$, the residual updater
$U_\psi$ predicts a correction conditioned on the selected object $o_{\pi_t}$,
step embedding $\mathbf{e}(s_t)$, and toolpath statistics
$\operatorname{stat}(\tau_t)$. These statistics summarize path extent, motion
modes, valid samples, and tool attributes. Adding the correction to $\bar{x}_t$
and clipping the result to $[0,1]$ yields $x_t$, which refreshes the MER tokens
used for subsequent operation and toolpath predictions. Recorded reference
states provide
supervision where available; final geometric quality is measured by the
offline material-removal evaluator.

\subsection{Verification-Guided Training}
\label{app:surrogate-verifier}

\subsubsection{Offline Material-Removal Labels}
Offline material-removal evaluation provides geometric labels for training
the surrogate verifier. The offline verifier constructs an occupancy grid
over the stock bounding box with a spacing of $4.0$\,mm. For synthetic
construction checks, target occupancy is determined by testing grid points
against the target B-rep. Real-data targets are constructed from manufacturing
annotations as described in Section~\ref{app:real-data-target-construction}.

To obtain machined occupancy, the verifier initializes the grid with stock
occupancy and removes the regions swept by the cutter during plunge and
cutting motions. For cylindrical stocks, the initial occupancy is additionally
restricted by the stock radius in the $xy$ plane.

Geometric targets use the metric definitions in
Appendix~\ref{app:verification-metrics}.

For training and ranking, nonnegative error ratios are converted to bounded
penalties $p(r)=r/(1+r)$. This preserves their ordering while reducing the
influence of large errors on regression. The selected-region quantities used
in the generator loss in Equation~\ref{eq:sv_loss} are SV predictions; the
reported benchmark geometry metrics come from offline evaluation.

\subsubsection{Surrogate Architecture}

The SV uses analytic path-contact estimates as inputs to a learned residual
calibration head. The analytic branch takes normalized waypoints, feature
geometry, valid-point masks, tool diameter, operation identity, and rollout
state. A smooth target signed-distance proxy distinguishes path length inside
and outside the intended feature. The resulting features provide initial
estimates of removal coverage, remaining target removal, overcut mass, aircut
mass, IoU, Removal F1, validity, and risk.

The calibration head combines these estimates with encoded material state,
object state, action embedding, and normalized step index. It predicts bounded
corrections for IoU, Removal F1, overcut penalty, and residual penalty, together
with logit corrections for validity and risk. The calibrated outputs are
clipped to their valid ranges.

After pretraining, the SV parameters are frozen during generator refinement.
Gradients propagate through the verifier outputs to the generator, providing
geometric feedback without updating the verifier parameters.

\subsubsection{Training Objective and Risk Labels}

Optimization settings for SV pretraining are provided in
Appendix~\ref{app:training-details}.

SV pretraining optimizes the verifier parameters using the refined-branch and
calibrated-branch losses:
\begin{equation}
\mathcal{L}_{\mathrm{SV\text{-}pretrain}}
=0.5\,\mathcal{L}_{\mathrm{refined}}+\mathcal{L}_{\mathrm{calibrated}}.
\end{equation}
This objective trains the verifier itself. During generator refinement, the
verifier is frozen and the separate objective $\mathcal{L}_{\mathrm{sv}}$ in
Equation~\ref{eq:sv_loss} supplies geometric feedback to the generator.

The refined branch uses Smooth-$L_1$ losses for IoU, Removal F1, overcut penalty,
and residual penalty, with respective weights of 1.0, 1.0, 2.0, and 0.5. An
additional penalty for underestimating overcut has weight 1.0. The calibrated
branch uses the same weights for these four geometric terms and adds binary
cross-entropy losses for validity and risk, with a risk weight of 1.5. Overcut
receives greater weight because subsequent operations cannot restore removed
target material.

A candidate receives a rejectable-risk label if any of three conditions holds:
its overcut penalty exceeds 0.12, its removal recall is below 0.90, or its
verification report is invalid. Validity and risk are auxiliary diagnostics
during generator refinement. The generator's geometric penalty instead uses
the selected-region IoU, overcut, and residual terms defined in the main text.

\section{Experimental Details and Additional Results}
\label{app:evaluation-protocol-details}

\subsection{Implementation and Training Settings}
\label{app:training-details}

Reported experimental results are averaged over three runs with different random seeds.

Models are implemented in PyTorch and trained on two NVIDIA RTX A6000 GPUs
with bfloat16 automatic mixed precision. Feature recognition and operation
planning are pretrained with Adam for 30 epochs, learning rate $10^{-3}$,
per-GPU batch size 16, and global batch size 32. Toolpath generation uses
AdamW for 50 epochs, learning rate $10^{-4}$, weight decay $10^{-4}$, and a
ReduceLROnPlateau scheduler.

SV pretraining uses Adam with learning rate $10^{-3}$, batch size 1024, at most
50 epochs, and early stopping with patience 8. Available split annotations
define the training and validation partitions; otherwise, a deterministic
validation partition is reserved from the verification-training records.

Joint refinement uses Adam with module-specific learning rates of
$5\times10^{-6}$ for feature recognition, $5\times10^{-5}$ for planning,
$2\times10^{-5}$ for toolpath generation, and $1\times10^{-5}$ for the
joint-refinement parameters. This stage uses global batch size 8 and runs
for 50 epochs. The generator uses its own rollouts, and the SV remains frozen.

\subsection{Evaluation Protocol and Metrics}
\label{app:unified-geometry-evaluator}

Compared methods share canonical geometry records, B-rep preprocessing, stock
coordinates, millimeter units, and target-construction rules. Each exports
predicted occupancy in the common stock frame.

\paragraph{Real-record target construction.}
\label{app:real-data-target-construction}
Real records are evaluated with the near-net-stock interface used in the main
comparison. Verified manufacturing annotations define the target removal
occupancy; auxiliary voxel records support consistency checks. The evaluator
reconstructs near-net stock occupancy from stock and toolpath metadata, then
subtracts the manufacturing-derived removal volume to obtain the target part.
Boundary-profile setup machining is treated as a deterministic setup operation
and excluded from the removal target when appropriate. Pockets, holes,
chamfers, and slant features remain the evaluated machining targets.

\paragraph{Geometric metrics.}
\label{app:verification-metrics}
Occupancy-based metrics use a grid spacing of $4.0$\,mm. Let $S$ denote the
initial stock occupancy, $T$ the target part occupancy, and $\hat T$ the
predicted machined occupancy. The target and predicted removal volumes are
$R=S\setminus T$ and $\hat R=S\setminus\hat T$, respectively. We measure
final-shape agreement with IoU and material-removal quality with removal
precision, recall, and F1:
\begin{equation}
\operatorname{IoU}=\frac{|\hat T\cap T|}{|\hat T\cup T|+\epsilon},
\end{equation}
\begin{equation}
P_{\mathrm{rem}}=\frac{|\hat R\cap R|}{|\hat R|+\epsilon},\qquad
R_{\mathrm{rem}}=\frac{|\hat R\cap R|}{|R|+\epsilon},\qquad
F1_{\mathrm{rem}}=\frac{2P_{\mathrm{rem}}R_{\mathrm{rem}}}
{P_{\mathrm{rem}}+R_{\mathrm{rem}}+\epsilon}.
\end{equation}
The overcut and residual ratios distinguish removal of intended part material
from incomplete removal of the target region:
\begin{equation}
r_{\mathrm{overcut}}=\frac{|\hat R\cap T|}{|T|+\epsilon},\qquad
r_{\mathrm{residual}}=\frac{|\hat T\cap R|}{|R|+\epsilon}.
\end{equation}
Here, $\epsilon$ stabilizes the denominators. Overcut is normalized by the
target part volume $|T|$ and measures removal of material that should remain.
Residual is normalized by the target removal volume $|R|$ and measures the
fraction of intended removal that remains uncut. CD uses 8192 uniformly
sampled points on predicted and target surfaces extracted at the same
occupancy resolution, with distances normalized by the stock bounding-box
diagonal.

\paragraph{Failed executions.}
Failed executions remain in aggregate evaluation. A partial rollout is scored
using its last exported occupancy. If a method exports no occupancy, its
prediction is the uncut stock. Such cases are also retained for qualitative
failure analysis.

\subsection{Baseline Adaptation}
\label{app:cncnet-mrp-variants}

The Machining Region Prior (MRP) adapts CNC-Net~\citep{yavartanoo2024cncnet}
to the B-rep-to-final-geometry evaluation interface through a rasterized support
mask and depth field. CNC-Net and \model share the B-rep parser, stock
definition, units, top-at-zero coordinate convention, feature scope, and target
construction. CNC-Net consumes the rasterized support and depth fields, whereas
\model uses B-rep graph features with object, operation, toolpath, and state
supervision. The adapter retains CNC-Net's carving model without adding MER,
prefix refinement, state-conditioned toolpath decoding, or the SV training loss.

\paragraph{Predicted region priors.}
CNC-Net+Pred.\ MRP predicts the prior from the input B-rep using a lightweight
region head trained on the synthetic training split. At test time, generation
uses predicted priors without access to reference manufacturing labels, target
occupancy, or target removal volume.

\paragraph{Oracle region priors.}
CNC-Net+Oracle MRP evaluates process generation with regions and depths supplied
by reference manufacturing-feature and toolpath records. The support channel
is the union of retained feature footprints, and the depth channel records
removal depth. Holes use circular footprints, pockets use polygonal footprints,
and chamfers and slant features use edge-band approximations. Overlapping
features use the maximum removal depth. Boundary-profile setup operations are
excluded when they are excluded from the evaluation target. The
annotation-derived prior makes this a privileged diagnostic setting.

\paragraph{Training and evaluation conditions.}
The two MRP variants share tensor format, rasterization resolution, input
channels, evaluator-grid mapping, data splits, and model-selection rules.
CNC-Net+Pred.\ MRP is trained and tested with predicted priors;
CNC-Net+Oracle MRP is trained, validated, and tested with oracle priors.
Each variant therefore retains its prior source between training and evaluation.

\subsection{Ablation Settings and Analysis}
\label{app:detailed-ablation-analysis}

The ablations in Table~\ref{tab:ablation} remove one component at a time under
the common evaluation protocol. Their joint interpretation concerns target
identity, sequential correction, state-conditioned motion, and geometric
training feedback.

\paragraph{Without MER.}
Removing MER eliminates stable operation-target identities, making the
object-referenced Step EM and Seq.\ EM metrics inapplicable. Category Macro-F1
falls from 0.9024 to 0.2182 and Removal F1 from 0.9016 to 0.1484. CD increases
from 0.0268 to 0.0317, while residual material rises from 0.0431 to 0.0673.
These differences support a role for object memory in coordinating categorical
predictions with material removal.

\paragraph{Without prefix refinement.}
This variant disables the prefix-refinement module and uses the base planner
to produce the operation sequence. Step EM decreases from 0.8552 to 0.8172
and Seq.\ EM from 0.8290 to 0.7796. Removal F1 falls to 0.3115, while overcut
increases from 0.0079 to 0.0167. These results support the contribution of
prefix refinement to operation-sequence accuracy and the resulting
material-removal quality.

\paragraph{Without state-aware toolpaths.}
This variant replaces the global material-state and object-state inputs to
the toolpath generator, including their grid representations, with zeros,
while retaining object features, the static stock description, and the
operation category. Step EM is 0.8530 and Seq.\ EM is 0.8260, close to the
full model's 0.8552 and 0.8290. However, Removal F1 falls from 0.9016 to
0.3113, IoU from 0.9525 to 0.9327, and overcut increases from 0.0079 to
0.0169. These results show that similar operation-sequence accuracy can
accompany substantially different material-removal quality, supporting
explicit state conditioning in toolpath generation.

\paragraph{Without the SV loss.}
This variant sets the weight of the surrogate-verifier loss to zero during
generator refinement, while retaining object memory, prefix refinement,
state-conditioned toolpath generation, and the remaining training losses.
Generated outcomes are still evaluated using the offline geometry evaluator.
IoU decreases from 0.9525 to 0.9296 and Removal F1 from 0.9016 to 0.3041.
CD rises from 0.0268 to 0.0286, while overcut reaches 0.0236, the highest
value among the ablations. These results support the contribution of
verification-guided training to material-removal quality, particularly in
reducing unintended removal of target material.

\subsection{Surrogate Verifier Reliability}
\label{app:sv-reliability}

We evaluate SV predictions against offline verification labels on held-out
records. For continuous targets, MAE and RMSE measure prediction error, while
Spearman correlation measures agreement in candidate ordering. For binary risk
and validity labels, we report AUROC, precision, recall, F1, and accuracy.
Tables~\ref{tab:sv-continuous-reliability} and~\ref{tab:sv-binary-reliability}
summarize these results.

\begin{table}[htbp]
    \centering
    \caption{Held-out SV agreement with continuous geometric targets.
    MAE and RMSE measure prediction error; Spearman correlation measures ranking agreement.}
    \label{tab:sv-continuous-reliability}
    \small
    \begin{tabular*}{\textwidth}{@{\extracolsep{\fill}}lccc@{}}
        \toprule
        Target & MAE $\downarrow$ & RMSE $\downarrow$ & Spearman $\rho$ $\uparrow$ \\
        \midrule
        IoU                 & 0.048 & 0.071 & 0.79 \\
        Target preservation & 0.021 & 0.035 & 0.61 \\
        Overcut penalty     & 0.006 & 0.019 & 0.76 \\
        Residual penalty    & 0.056 & 0.083 & 0.74 \\
        \bottomrule
    \end{tabular*}
\end{table}

\begin{table}[htbp]
    \centering
    \caption{Held-out SV risk and validity screening.}
    \label{tab:sv-binary-reliability}
    \small
    \begin{tabular*}{\textwidth}{@{\extracolsep{\fill}}lccccc@{}}
        \toprule
        Target & AUROC $\uparrow$ & Precision $\uparrow$ & Recall $\uparrow$ & F1 $\uparrow$ & Acc. $\uparrow$ \\
        \midrule
        Risk label     & 0.958 & 0.902 & 0.887 & 0.894 & 0.901 \\
        Validity label & 0.951 & 0.889 & 0.932 & 0.910 & 0.904 \\
        \bottomrule
    \end{tabular*}
\end{table}

Spearman correlations of 0.61--0.79 indicate positive agreement between SV
predictions and offline target rankings. The risk and validity classifiers
achieve AUROC values of 0.958 and 0.951, with corresponding F1 scores of 0.894
and 0.910. These diagnostics quantify the surrogate's prediction agreement and
binary screening performance on held-out records. Final machining geometry is
evaluated separately using the offline material-removal evaluator.

\subsection{Additional Qualitative Results and Failure Analysis}
\label{app:failure-case-analysis}

\begin{figure}[htbp]
    \centering
    \includegraphics[width=\textwidth,height=0.58\textheight,keepaspectratio,trim=18pt 18pt 18pt 18pt,clip]{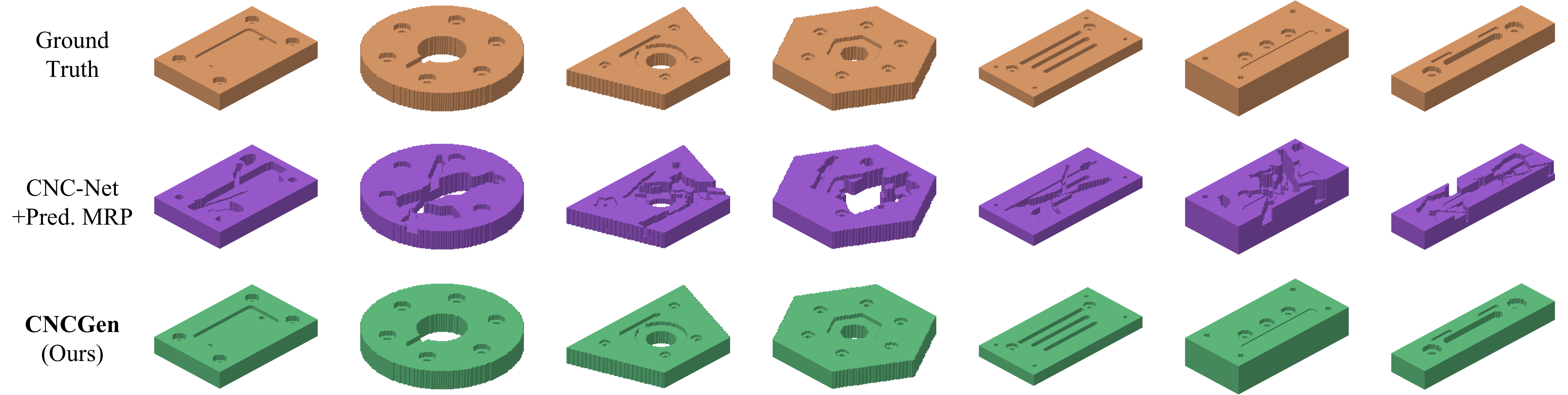}
    \caption{Geometric outcomes on held-out real CNC parts. Rows show target
    geometry, CNC-Net+Pred.\ MRP, and \model, from top to bottom, under the
    common offline evaluator.}
    \label{fig:supp-realcompare}
\end{figure}

\paragraph{Held-out real parts.}
Fig.~\ref{fig:supp-realcompare} provides an enlarged view of the qualitative
comparison summarized in Fig.~\ref{fig:realcompare} of the main text.

In these examples, CNC-Net+Pred.\ MRP recovers coarse removal regions but
leaves fragmented residual structures or incomplete feature interiors.
\model more completely recovers the displayed cavities and through-regions.
The quantitative comparison in Table~\ref{tab:realcnc-verification-generalization}
reports the corresponding differences in removal completeness and geometric error.

\begin{figure}[htbp]
    \centering
    \includegraphics[width=\textwidth,height=0.40\textheight,keepaspectratio,trim=18pt 18pt 18pt 18pt,clip]{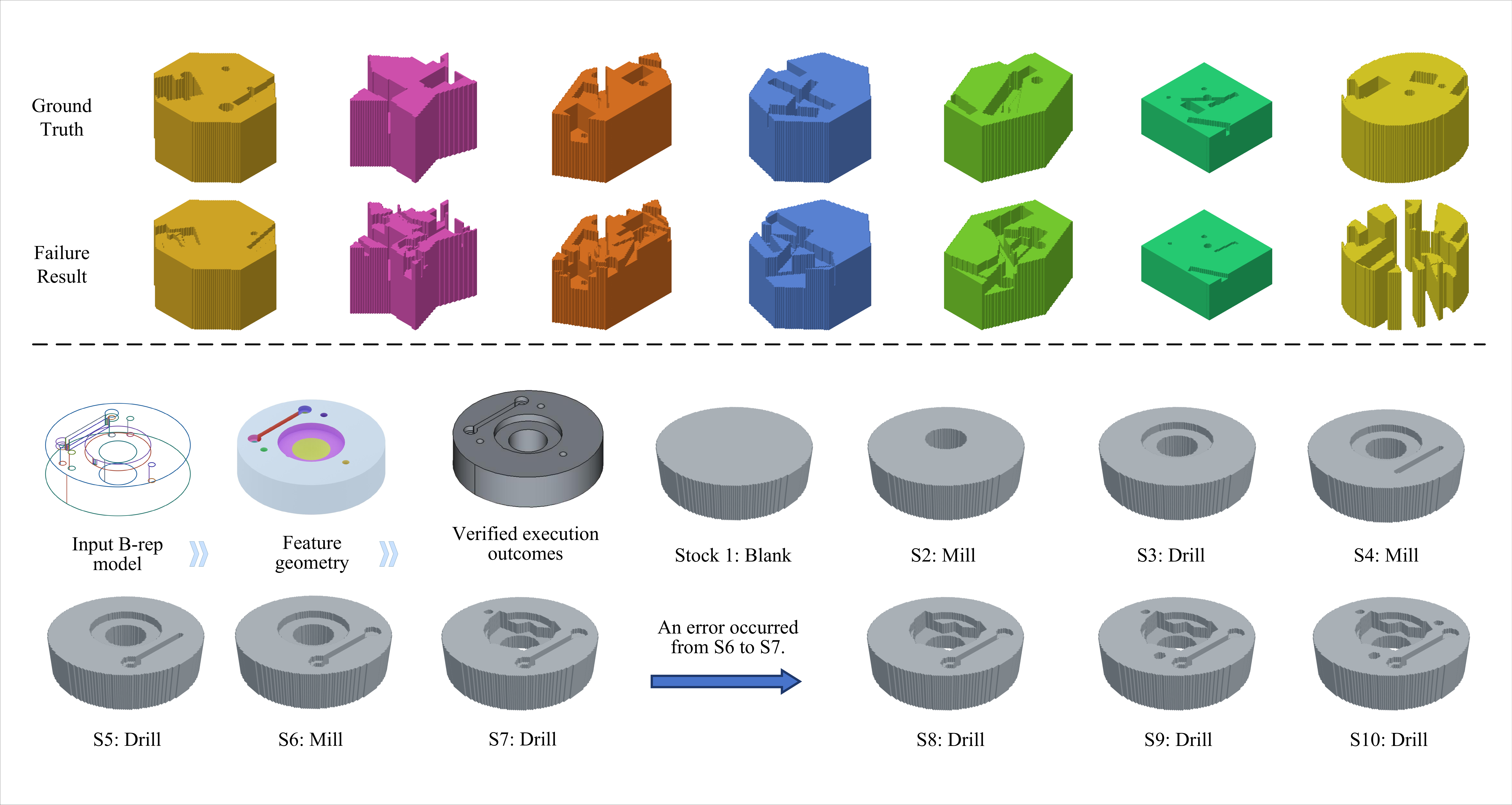}
    \caption{Representative failed rollouts. Target geometries in the top row
    are compared with generated results in the bottom row, showing residual
    material, overcut, and errors near feature boundaries.}
    \label{fig:supp-failure-cases}
\end{figure}

\paragraph{Failure patterns.}
Fig.~\ref{fig:supp-failure-cases} shows rollouts with low executed-geometry
scores. Incomplete feature recovery leaves residual material near narrow
bands, pocket intersections, and slant-feature boundaries, even when the main
pocket or hole region has been removed. Residual material is therefore useful
for inspecting local errors that can remain within an otherwise accurate
final shape.

Other failures contain fragmented or noisy geometry outside the intended
feature footprint. Case inspection associates these patterns with incorrect
object binding or inaccurate local paths after accumulated state errors.
They reduce IoU and increase overcut even when the predicted operation
category is reasonable. Narrow chamfers, slant features, and overlapping
pocket boundaries are particularly sensitive to local path errors. These
examples identify cases in which compact state tracking and B\'ezier
parameterization leave small regions insufficiently corrected at later steps.

\section{Extended Discussion}
\label{app:limitations-details}

\paragraph{Process and motion coverage.}
The current representation covers common three-axis pockets, holes, chamfers,
and slant features. External profile machining is handled through setup
construction rather than learned as an explicit target. Turning, grinding,
EDM, thread milling, free-form finishing, and simultaneous multi-axis
machining require broader process and motion representations.

\paragraph{Physical execution.}
Generated plans and parameterized toolpaths are controller-neutral.
Machine-specific post-processing must provide the corresponding control
program. Geometric verification measures material removal; it does not
evaluate cutting force, tool deflection, chatter, thermal effects, tool wear,
coolant behavior, or the physical suitability of predicted feeds and speeds.
Fixture-aware collision checking and physical process models would extend
the conditions assessed during generation and evaluation.

\paragraph{Data coverage and interacting features.}
Synthetic training provides controlled coverage of standard-part families,
while the 800 real test records cover a limited range of shop practices,
fixtures, and machining intent. Broader real records would support evaluation
of these variations. The observed boundary-sensitive failures also motivate
richer local state representations and more precise late-stage path
correction for interacting features.

\end{document}